\documentclass{easychair}
\usepackage{listings}
\usepackage{graphicx}
\usepackage{multirow}
\title{GpuShareSat: a SAT solver using the GPU for clause sharing}
\titlerunning{GpuShareSat}

\institute{} 

\author{Nicolas Prevot \\
  \href{mailto:nicolas.prevt@gmail.com}{nicolas.prevt@gmail.com} }
\authorrunning{NicolasPrevot}

\usepackage[latin1]{inputenc}

\usepackage{amsopn}
\usepackage{amsmath}
\usepackage{amsfonts}
\usepackage{graphicx}
\usepackage{pgf}
\usepackage{tikz}
\usepackage{url}
\newtheorem{theorem}{Theorem}
\newtheorem{definition}{Definition}
\newtheorem{proposition}{Proposition}
\newtheorem{example}{Example}

\begin{document}
% \email{nicolasprevot2@gmail.com}
\maketitle

\begin{abstract}
We describe a SAT solver using both the GPU (CUDA) and the CPU with a new clause exchange strategy. The CPU runs a classic multithreaded CDCL SAT solver. Each CPU thread exports all the clauses it learns to the GPU. The GPU makes a heavy usage of bitwise operations. It notices when a clause would have been used by a CPU thread and notifies that thread, in which case it imports that clause. This relies on the GPU repeatedly testing millions of clauses against hundreds of assignments.
All the clauses are tested independantly from each other (which allows the GPU massively parallel approach), but against all the assignments at once, using bitwise operations.
This allows CPU threads to only import clauses which would have been useful for them.

Our solver is based upon glucose-syrup.
Experiments show that this leads to a strong performance improvement, with 22 more instances solved on the SAT 2020 competition than glucose-syrup.
\end{abstract}

\section{Introduction}
Boolean satisfiability (SAT) is a fundamental problem in computer science. Despite being NP complete, modern SAT solvers are able to handle large instances with millions of variables. It is widely used for a variety of application, which includes theorem proving and hardware verification. 

Graphical Computing Unit (GPU) are highly parallel devices, and are able to perform many times more operations per second than the CPU.
CUDA, provided by NVIDIA, is a popular programmable platform to perform general purpose computation on the GPU. However, transposing SAT algorithms from the CPU to the GPU is not an easy task.

A problem for multithreaded SAT solvers is to decide which clauses to share between threads and which ones not to. We present a novel algorithm which relies on the GPU via CUDA to tackle this problem. Our solver builds upon glucose-syrup 4.1, a multithreaded CPU SAT solver which is itself based upon MiniSat. The code for the CDCL procedure of CPU solver threads is mostly kept. The code to exchange clauses between them is not.

\section{Related work}

\subsection{Sat solving and the GPU}
\label{ssec:workGpu}
Modern SAT solvers algorithm (like the DPLL procedure) which are effective on the CPU do not translate easily to the GPU. One difficulty is that alghough the GPU can run many times more threads at once than the CPU, the amount of memory and cache available per thread is much smaller on the GPU. Running the DPLL procedure with the watched literal scheme requires a sizeable amount of memory, especially on large formulas. This would make running the DPLL procedure separately in each GPU thread challenging. Alternatively, using several GPU threads to do propagations for a single DPLL procedure would requires heavy communication between several GPU threads, which is also difficult.
Various approaches have been proposed that use the GPU to solve SAT but they have not become widespread. In ~\cite{cud@sat}, the GPU is used to do unit propagation. 
In ~\cite{OsamaGPUSimp}, the GPU is used for pre-processing the formula (variable elimination, subsumption)... 
In ~\cite{SurveyGpu}, the GPU is used for the survey propagation algorithm.

However, most of the modern SAT solvers rely exclusively on the CPU.

\subsection{Clause exchange in CPU multithreaded SAT solvers}
In a multithreaded SAT solver: on the CPU, each thread runs the DPLL procedure separately from the others and learns clauses.
For threads to benefit from each other's work, they may share some clauses with each other. 
One difficulty with this approach is to decide which clauses should be shared with the other threads and which should not. Sharing too many clauses would increase the number of clauses that each thread has, which slows down the search. However, if not enough clauses are shared, each thread does not benefit for the work of other threads.
So a heuristic is used to identify which clause is good and should be shared.
In ~\cite{AudemardExchangeParallel}, clauses are shared only if their size (or lbd) is smaller than a constant.
~\cite{vallade2020community} uses a metric based on the community structure.
In ~\cite{audemard2014lazy}, a thread will only share a clause if it has been used another time.
In ~\cite{EhlersMassivelyParallel}, a thread will only export a clause to another thread if they are connected by an edge in a graph.

\section{A new clause exchange strategy}

\subsection{Preliminaries}
We define a truth value as a member of the set \{T, F, U\} where T stands for True, F for False, and U for Undef \\
The negation of a truth value w is denoted as $\neg w$, with $\neg T = F$, $\neg F = T$, $\neg U = U$ \\
An assignment A is a function from a set of variables to a truth value\\
A literal l is a variable v or its negation $\neg v$\\
Given an assignment A and a literal l, $A(l) = A(v)$ if $l = v$ and $A(l) = \neg A(v)$ if $l = \neg v$\\
A clause C of size s is a disjunction of n literals: $C_1 \lor C_2$...$\lor C_s$\\
A formula is a conjunction of clauses\\

\subsection{Clauses used recently are likely to be used again soon}
SAT solvers aim at keeping useful clauses, and deleting bad ones.
Two different ways to measure a clause usefulness have been described in e.g. ~\cite{predictingClauseQuality} : we can measure how often a clause is used during conflict analysis, or during unit propagation (how often it implies a previously unset variable).  
We wanted to investigate how likely it for a clause which has been used recently to be used again soon. 

\begin{figure}[h]
\caption{Distribution of conflict intervals}
\label{fig:distribution_conflict_interval}
\includegraphics[width=\textwidth]{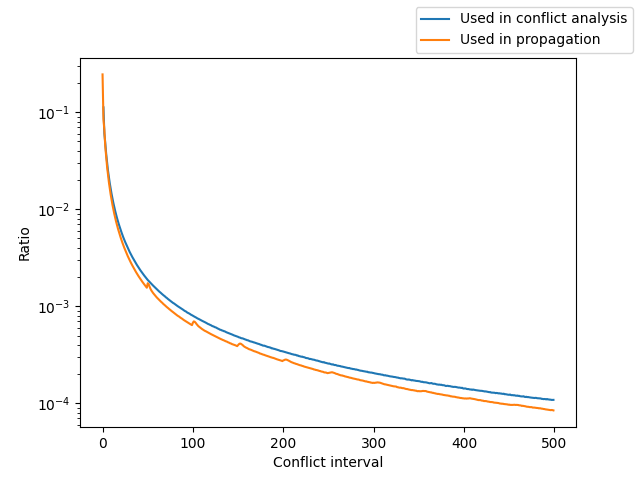}
\end{figure}

On 10 instances taken at random from the SAT 2020 competition main track:
whenever a clause was used, we recorded the conflict interval with the last time it was used, that is, how many conflicts ago it was last used. We did this for usage in both unit propagation and conflict analysis. Whenever a clause was used for the first time, we considered the conflict interval to be infinite.
From \ref{fig:distribution_conflict_interval}, the longer the conflict interval, the less frequent it is.

\begin{figure}[h]
\caption{Cumulative distribution of conflict intervals}
\label{fig:cum_distribution_conflict_interval}
\includegraphics[width=\textwidth]{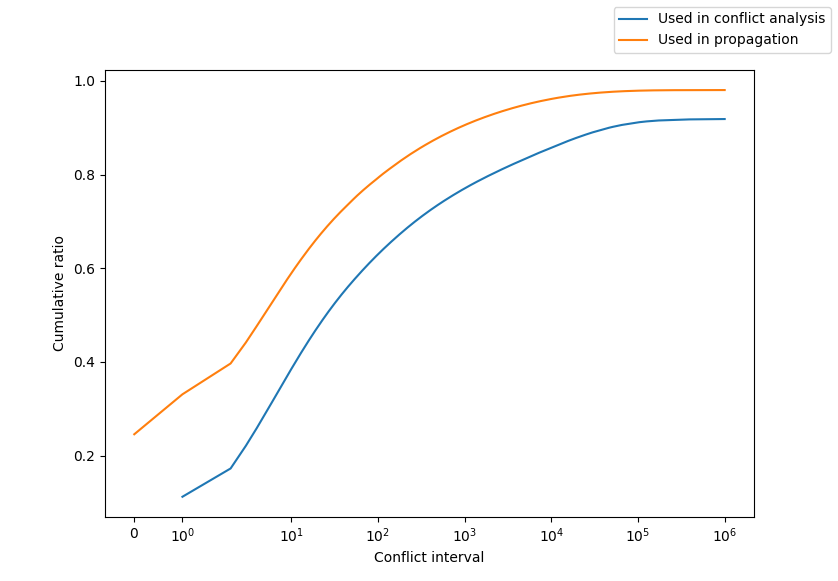}
\end{figure}

Let's consider a clause which has just been used. The distribution of conflict intervals indicates after how many conflicts it will be used again.
From \ref{fig:cum_distribution_conflict_interval}, a clause which has just been used in conflict analysis has a 39\% chance to be used again within the next 10 conflicts, and a 62 \% chance to be used again within the next 100 conflicts.
This tells us that \textit {clauses used recently are likely to be used again soon}

\subsection{Importing clauses which would have been used}
Let's consider consider how this applies to the problem of importing clauses learned by other threads.

Consider an assignment of a thread where unit propagation completed without conflict, and a clause which this thread does not have (for example because it was learned by another thread).
If all literals of this clause are False except for one which is Undef, then this clause would have implied this Undef literal, so it would have been used in unit propagation.
If all literals of this clause are False, then this clause would have been in conflict, so it would have been used in conflict analysis.
This allows us to identify some clauses which would have been used.

From the previous section, clauses used recently are likely to be used again soon. We can extrapolate this to the following: \textit {If a thread imports a clause which would have been used recently, it is likely to be used soon}.

This forms our importing strategy: \textit {A thread should import clauses which would have been used recently.}
We identify cases where a thread did not have a clause, but it would have been used if it did have it. In this case, this clause is likey to be used again, so the thread will import it. This will be the only mechanism we will use for a thread to import clauses learned by other threads.

\begin{definition}
A clause C of size s triggers on an assignment A if:
\begin{itemize}
\item For each literal l in C, $A(l) \neq T $ 
\item For at least s - 1 literals l in C, $A(l) = F$ 
\end{itemize}
\end{definition}

\begin{example}
Given the assignment $(A \leftarrow F, B \leftarrow F, C \leftarrow F, D \leftarrow U)$, the clause $A \lor B \lor D$ triggers (it would have implied $D \leftarrow T$). The clause $A \lor C$ triggers too (it would have been in conflict). The clause $A \lor \neg B$ does not.
\end{example}

A clause which triggers on an assignment of a thread where unit propagation completed without conflict would have been used by this thread. 

To find clauses which trigger and thus would have been used: rather than considering all assignments where unit propagation completed without conflict, we will limit ourselves to the parents of the conflicts: that is, the last assignment where unit propagation completed without conflict before a conflict.
This is simply to test fewer assignments.

\subsection {Testing which clauses trigger}
As described in ~\cite{BitLevel}, we can use bitwise operations to test if a clause triggers on multiple assignments at once.
Given N assignment $A_i$ for $1 \leq i \leq N$ and a variable v:
we can effectively represent the values $A_i(v)$ using only two N-bits variables: (isTrue, isSet).\\
The i-th bit is isSet represents whether $A_i(v) \neq U$\\
The i-th bit is isTrue is set if $A_i(v) = T$, is not set if $A_i(v) = F$ and might be set or not if $A_i(v) = U$\\
Given a literal $l = \neg v$, the values $A_i(l)$ can be computed with $(\sim isTrue, isSet)$\
This allows us to test if a clause C of size s triggers over N assignments at once using bitwise operations:

\begin{minipage}{\linewidth}
\begin{lstlisting}[label={lst:assigTrigger}]
assignmentTrigger(B bitwise assignments, C clause)
    allFalse <- ~0
    oneUndef <- 0
    for l in C:
        oneUndef = (allFalse & ~B.isSet(l)) | (oneUndef & B.isFalse(l))
        allFalse &= B.isFalse(l)
    return allFalse | oneUndef
\end{lstlisting}
\end{minipage}
The previous algorithm returns an N bits variables, where the bit i is set if the clause C triggers on the assignment i.

\subsection{GPU implementation}
Given a set of up to 32 assignments, and a set of clauses (whose number can reach several millions):
We test in parallel if each of the clauses triggers on these assignments (i.e. is in conflict or is implying something currently Undef). This approach fits very well with the GPU since it is massively parallelisable.
We are doing two forms of parallelism: 
\begin{itemize}
\item Many GPU threads run at the same time and test different clauses
\item Each GPU thread tests multiple assignments at once with bitwise operations
\end{itemize}

To make reading the clauses on the GPU faster, we also coalesce reads to the clauses by reordering how we represent them in memory.

\subsection{Modifications to the CPU part of a SAT solver}
Like in traditional parallel SAT solvers, GpuShareSat relies on having several CPU threads, each one running the CDCL procedure in a separate solver.

In GpuShareSat, each CPU solver thread exports all clauses it learns to the GPU (not directly to other CPU threads).
Whenever a CPU thread reaches a conflict, it sends the assignment of the parent of the conflict to the GPU. 
During a GPU run, the GPU tests all of the clauses it has against all the assignments sent by the CPU threads it has not tested yet. It reports all the clauses that trigger.
Unfortunately, it is possible that the GPU is not be able to cope with all the assignments coming from the CPU. This is because the GPU can only test a finite number of assignments during a run. In this case, some of these assignments will not be sent to the GPU.
The CPU threads import the clauses reported to them. These clauses would have been used recently, so are likely to be used again soon. So, we want to import them immediately rather than wait until we are back at level 0.

This required a modification of the code which imports clauses to be able to import clauses at any time, and keep the watched literal scheme consistent:
\begin{itemize}
\item If at least 2 literals of the imported clause are Undef or True, the clause can directly be attached.
\item If all literals are False except for one which is Undef: we take the highest level of the False literals, backtrack to that level, and imply the Undef literal there.
\item If all literals are False except for one which is True: we take the highest level of the False literals, call it l. If the level of the True literal is strictly higher than l, we backtrack to l and imply it there. Otherwise, we do nothing.
\item If all literals are False: let's call $l_1$ and $l_2$ the highest levels of literals in C with $l_1 \leq l_2$. if $l_1 < l_2$, we backtrack to l1 and imply the last literal there. Otherwise, we do conflict analysis at l1.
\end{itemize}

The (very simplified) algorithm of a CPU thread becomes:

\begin{minipage}{\linewidth}
\begin{lstlisting}
while noOtherThreadHasFoundAnAnswer():
    importAllClausesReportedByTheGpuToUs()
    clauseInConflict = propagate()
    if clauseInConflict is set:
        sendParentAssignmentToGpu()
        learnedClause = learnClauseFromConflictAnalysis()
        if level == 0:
            return Unsatisfiable
        else:
            sendClauseToGpu(learnedClause)
        backtrack
    else:
        if allVariablesAreSet():
            return Satisfiable
        makeDecision()
    maybeReduceClauseDb()
    maybeRestart()
\end{lstlisting}
\end{minipage}

\begin{example}
Let's assume that a CPU thread has the clauses $\neg A \lor B$, $\neg B \lor \neg D \lor \neg E$, $\neg B \lor \neg D \lor E$, and the GPU has $\neg A \lor \neg B \lor C$.
Assume the threads takes the decision $A \leftarrow T$. Thanks to $\neg A \lor B$ it will enqueue $B \leftarrow T$ and unit propation will stop there. \\
Assume it then takes the decision $D \leftarrow T$. Thanks to $\neg B \lor \neg D \lor \neg E$, it will enqueue $E \leftarrow F$. The clause $\neg B \lor \neg D \lor E$ will then be in conflict, so the thread will do conflict analysis and learn a clause. In addition, it will then send the parent assignment ($A \leftarrow T, B \leftarrow T$) to the GPU. The GPU will notice that $\neg A \lor \neg B \lor C$ triggers on this assignment (it would imply $C \leftarrow T$), and the CPU thread will import this clause.
\end{example}

\subsection{Why it is effective}
In a state-of-the-art parallel SAT solver, a thread may decide to export a clause it learns to other CPU threads.
As discussed previously, the difficulty with this approach is to decide which clauses to export. Our approach allows a thread to only import a clause (coming from the GPU) if it would have been useful recently.
This has the following benefits:
\begin{itemize}
  \item As seen previously, if a clause would have been used recently by a thread, it is likely to be used again.
  \item Clauses that would have been useful are probably better than those that wouldn't, so threads tend to import good rather than bad clauses
  \item This clause may still be useful at the time the CPU thread imports it. Unfortunately, in practice, this happens for less than 1\% of reported clauses
  \item If a CPU thread has deleted a clause as part of its clause deletion policy, the GPU may later notify it that this clause would have been useful, in which case it will re-import the clause. This contrast with state of the art SAT solvers where a clause deleted by a thread will be lost forever.
\end{itemize}

The GPU suffers from the same problem as the CPU solvers in that it also needs to delete clauses. Otherwise, it would become slow and run out of memory.
To do that, we perform activity based clause deletion, in a way similar to MiniSat on the CPU.
In our case, the activity of a clause is bumped whenever it triggers.

\section{Use of aggregate operations to boost the number of assignments}
\subsection{Commonalities between successive runs}
We wanted to know how different the values of variables in successive assignments coming from the same CPU thread are. 
So we devised an algorithm based upon single threaded glucose. In between 32 conflicts, for each variable, we computes all the values taken. We only consider the values of variables after backtracking, or in assignments where unit propagation has completed without a conflict. This gives us a subset of $\{T, F, U\}$ i.e. True, False, Undef. We then compute the ratio of the number of times each such subset happens over the entire run.

We ran it over 10 instances taken at random from the SAT 2020 competition. The average results follow:
\begin{center}
\begin{tabular}{ |c|c| }
\hline
subset & ratio \\
\hline
\{T\} & 0.127 \\
\{F\} & 0.064 \\
\{T, F\} & 0.000 \\
\{U\} & 0.660 \\
\{T, U\} & 0.060 \\
\{F, U\} & 0.068 \\
\{T, F, U\}  & 0.021 \\
\hline
\end{tabular}
\end{center}
We see that successive assignments coming from the same CPU thread have a good amount in common.
This is helpful when testing a clause againsts these successive assignments at once.
If a clause has the literals l1 l2 and the values taken for them does not include False: then we are sure that the clause will not trigger for any of these successive assignments.
This is because for a clause to trigger, it needs all literal values except for one to be False.
Alternatively, if there is a single literal among the clause for which the only value taken is True, then again, the clause will not trigger for any of these successive assignments.

So, given 32 assignments: by only looking at these three bits for a given variable: T, F, U, we are, in some cases, able to tell that the clause does not trigger.
The remaining of this section will formalise this and use it to test up to 32 times 32 assignments at once.

\subsection{Definitions and theorem}
\begin{definition}
An aggregate g is a subset of truth values ie $\{T, F, U\}$
\end{definition}

\begin{definition}
Given an aggregate g, we denote by $\neg g$ the aggregate $\{\neg w\}$ for $w \in g$
\end{definition}

An aggregate g can be efficiently represented using only three booleans: $(T \in g, F \in g, U \in g)$

\begin{definition}
Given a set of variables, An aggregate assignment G is a function which maps a variable to an aggregate
\end{definition}

\begin{definition}
Given an aggregate assignment G and a literal l, we define $G(l)$ as $G(v)$ if $l = v$ and 
$\neg G(v)$ if $l = \neg v$
\end{definition}

\begin{definition}
Given the assignments $(A_i)_{1 \leq i \leq N}$ , their associated aggregate assignment G is defined by $G(v) = \{A_i(v)\}_{1 \leq N}$. 
\end{definition}
As seen previously, N truth values can be represented by the two N-bits variables: isTrue, isSet\\
Their associated aggregate can be computed using only bitwise operations by: $ (isTrue \neq 0, (isSet \& \sim isTrue) \neq 0, (\sim isSet) \neq 0) $

\begin{example}
Given the assignments $(A \leftarrow T, B \leftarrow F, C \leftarrow U)$ and $(A \leftarrow T, B \leftarrow U, C \leftarrow T)$, their associated aggregate assignment is $(A \leftarrow \{T\}, B \leftarrow \{F, U\}, C \leftarrow \{U, T\})$
\end{example}

\begin{proposition}
Given the assignments $(A_i)_{1 \leq i \leq N}$, G their associated aggregate assignment, and a literal l: $G(l) = \{A_i(l)\}$
\end{proposition}

Demonstration:
This is true by definition if l = v\\
If $\l = \neg v$, then: \\
$G(l) = \neg G(v) = \neg \{A_i(v)\}_{1 \leq i \leq N} = \{\neg A_i(v)\}_{1 \leq i \leq N} = \{A_i(l)\}_{1 \leq i \leq N} $

\begin{definition}
A clause C of size s triggers on an aggregate assignment G if:
\begin{itemize}
\item for every literal l in C, $F \in G(l)$ or $U \in G(l)$
\item for at least s - 1 literals l in C, $F \in G(l)$
\end{itemize}
\end{definition}

\begin{theorem}
Given a set of assignments and their associated aggregate assignment: if a clause triggers on an assignment, it triggers on the aggregate assignment
\end{theorem}

Demonstration:
Given the assignments $(A_i)_{1 \leq i \leq N}$ and their associated aggregate assignment G: this is obvious using the definitions of a clause triggering and the fact that $A_i(l) \in G(l)$

By contraposition of this theorem: if a clause does not trigger on an aggregate assignment, it does not trigger on any individual assignment.

\begin{example}
Given the assignments $(A \leftarrow F, B \leftarrow F, C \leftarrow U)$ and $(A \leftarrow T, B \leftarrow F, C \leftarrow U)$,
The clause $A \lor B \lor C$ triggers on the first (it would imply $C \leftarrow T$) but not the second. 
It triggers on their associated aggregate assignment $(A \leftarrow \{T, F\}, B \leftarrow \{F\}, C \leftarrow \{U\})$
\end{example}

\begin{example}\label{ex:noTriggerIndividual}
Given the assignments $(A \leftarrow F, B \leftarrow T$ and $(A \leftarrow T, B \leftarrow F$,
The clause $A \lor B$ does not trigger on either of them. 
It does trigger on their associated aggregate assignment $(A \leftarrow \{T, F\}, B \leftarrow \{T, F\})$ though. This example proves that a clause may trigger on an aggregate assignment without triggering on any individual assignment.

\end{example}

\subsection{Usage in GpuShareSat}
We previously proposed an algorithm which returned whether a clause triggers over any of N assignments (with $N \leq 32$). We are going to propose a new algorithm which returns whether a clause triggers on a much larger number of assignments

This matters because when the number of CPU solver thread increases, it is less likely that the GPU is going to be able to keep up with the assignments coming from them.

We have a clause, M threads, and up to N assignments per thread (so up to MxN assignments in total). We want to find on which assignments the trigger test for this clause is positive. We could use \ref{lst:assigTrigger} M times on up to N assignments each time. But we can do better by using aggregate operations.
From the previous section, we can test a clause on a group of assignments by computing their aggregate assignment. We can borrow an idea from group testing: we can start by testing the clause on assignment groups, and only test it on the assignments within a group if the group tested positive.
From \ref{ex:noTriggerIndividual}, a clause may test positive on an assignment group without testing positive on any assignment within this group. 
This is especially likely if the assignments within the group are very different from each other. To reduce the number of tests, we want to avoid this. As seen previously, successive assignments coming from a single CPU solver thread are similar to each other. So, we will only group together assignments coming from the same CPU thread.

We saw in \ref{lst:assigTrigger} that we could test 32 individual assignments at once.
Similarly, we can test 32 assignment groups at once via their aggregate.
If we have M assignment groups: for each group, we compute its aggregate assignment.
Each aggregate assignment $G_i$ has three booleans for each variable: $(t_i(v), f_i(v), u_i(v))$
We can represent the M aggregate assignments using three M-bit variables: \[(canBeTrue(v), canBeFalse(v), canBeUndef(v))\]
Where the bit i of $canBeTrue(v)$ represents $t_i(v)$ and respectively for $canBeFalse(v)$ and $canBeUndef(f)$

The following algorithm returns on which aggregate assignments a clause C triggers:

\begin{minipage}{\linewidth}
\begin{lstlisting}
aggregateTrigger(G bitwise aggregate assignments, C clause)
    allFalse <- ~0
    oneUndef <- 0
    for l in C:
        oneUndef = (allFalse & G.canBeUndef(l)) | (oneUndef & G.canBeFalse(l))
        allFalse &= canBeFalse(l)
    return allFalse | oneUndef
\end{lstlisting}
\end{minipage}

It returns an M bit variable whose bit i is set if C triggers on the aggregate assignment $G_i$

As previously demonstrated, if a clause does not trigger on an aggregate assignment, it does not trigger on any individual assignment. Therefore, we can run the algorithm above first, and only test if the clause triggers on the individual assignments if it did not on their aggregate assignment.

\begin{minipage}{\linewidth}
\begin{lstlisting}[label={lst:multiTrigger}]
multiTrigger(G bitwise aggregate assignments, A assignments, C clause)
    agTrigger = aggregateTrigger(G, C)
    for bit i set in agTrigger:
        assigTrigger = assignmentTrigger(A[i], C)
        if assigTrigger != 0:
            report(i, assigTrigger)
\end{lstlisting}
\end{minipage}
The previous algorithm calls report(i, assigTrigger) whenever C triggers on some assignments of assignment group i. The bit j of assigTrigger is set if C triggers on the assignment j of assignment group i. We will then find the CPU solver thread which owns the assignment group i, and report this clause to it.

\begin{figure}[h]
\caption{Example trigger tests on assignments and aggregate assignments. The positive tests are in blue. The negative ones are in red. Those that did not need to be tested are in grey}
\includegraphics[width=\textwidth]{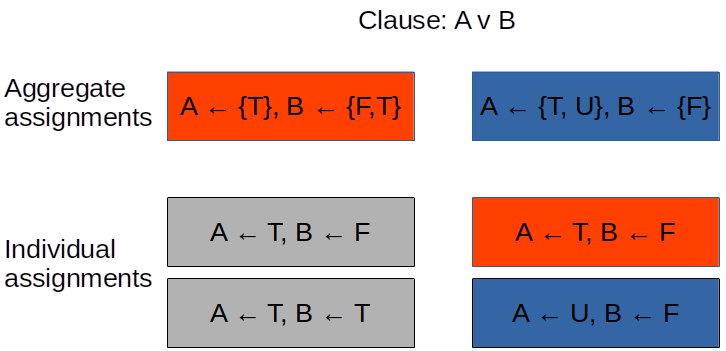}
\end{figure}

\section{Results}
We compared GpuShareSat(GPU and CPU) to glucose-syrup 4.1(CPU only) on the 2020 SAT competition main track with a timeout of 5000 seconds.
For the CPU hardware, we used a i9-9960x processor, 3.1 Ghz, with a processor cache of 22MB, 16 cores. 64 GB RAM.
For the GPU, an NVIDIA Quadro P6000 with 24GB memory was used.
32 CPU solver threads were used in glucose-syrup, and 31 in GpuShareSat (since there is another thread to manage the interaction with the GPU)

\begin{center}
\begin{tabular}{ |c|c|c| }
\hline
 & glucose-syrup & GpuShareSat\\
\hline
Total solved  & 250 & 272\\
Sat solved  & 123 & 138\\
Unsat solved  & 127 & 134\\
Score  & 1659500 & 1432316\\
\hline
\end{tabular}
\end{center}

\begin{figure}[h]
\caption{Solved instances by running time}
\includegraphics[width=\textwidth]{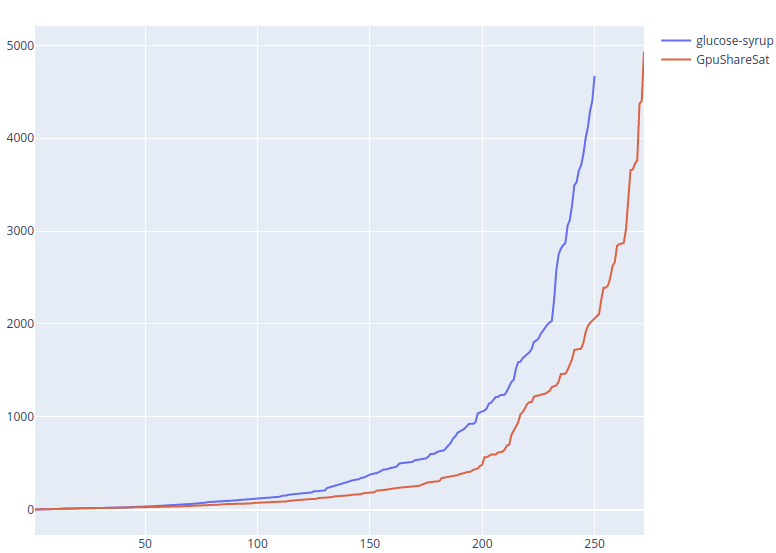}
\end{figure}

For each instance of the SAT 2020 competition main track, we computed the number of clauses tested by the GPU per second (for all CPU solver threads). The average value (weighted by the time spent on this instance) was 594 millions per second.

Similarly, we computed the ratio of assignments that CPU solver threads were unable to send to the GPU, due to it being already full. The average result was 0.103

Similarly, we computed the ratio of aggregate tests that were negative (which concluded that the clause does not trigger on any individual assignment). The result was 0.999422.
As a reminder, for a GPU run, the assignments sent by a CPU solver thread will be divided into one or more assignment groups.
This result is not so surprising considering that the learned clauses are often long (20 or more literals). As seen previously, we only need two literals of a clause to be Undef on all assignments of an assignment group for the aggregate test to be negative and conclude that the clause does not trigger on any assignment.

Similarly we computed the average number of clauses imported for each assignment sent to the GPU, the result was 2.132.
Similarly, the average number of clauses present on the GPU at the end of the run was 5.05 millions.

\section{Comparison with watched literal scheme}
\subsection{The watched literal scheme}
The two watched literal scheme ~\cite{EhlersMassivelyParallel} is used by modern CPU SAT solvers to tell when a clause is in conflict or implying some variable.
For each clause, it keeps two literals such that either:
\begin{itemize}
\item The value of one of them is True. If the other one is False, then its level must be greater or equal to the level of the True one.
\item The value of both of them is Undef.
\end{itemize}
Whenever a watched literal becomes False (and the other one is not True), the solver will iterate over the clause to find another literal with value True or Undef to watch. If no such literal can be found, the clause is either in conflict, or implying a variable.

\subsection{Comparison}
The algorithm code~\ref{lst:multiTrigger} is used to tell which clauses trigger on some assignments, just like the watched literal scheme. The watched literal scheme has much lower latency, it is much faster at noticing that a clause triggers. It will notice all clauses that trigger before doing another decision. In comparison, our GPU based algorithm takes more time to notice which clause triggers. 
It is heavily parallelizable and able to run on the GPU, though.

\subsection{Efficiency of GPU trigger tests}
Given a set of successive assignments coming from a single CPU solver thread: we are going to show that if a clause triggers on their aggregate assignment: if we had applied the two-watched literal scheme on this same clause: it would have had to look at the clause.

Suppose that the clause C of size s triggers on the aggregate assignment. On the first assignment, let's call l1 and l2 the two literals it watches. Let's also assume that the clause does not trigger on this assignment.
By definition of the aggregate assignment, there are at least s - 1 literals in C for which at least one assignment takes the value False.
So at least one of l1 or l2 takes the value False for at least one assignment.
So the watched literal scheme would have had to look at this clause at least once.

As a note, we do not send all assignments coming from a CPU solver thread to the GPU, only some of them. But the above stays true.

Let's consider the algorithm code~\ref{lst:multiTrigger}, a clause and a CPU solver thread.
The number of calls to assignmentTrigger for this clause and the assignments of this solver is lower than the number of times we would have looked at this clause as part of the two-watched literal scheme if this clause was attached to this CPU solver thread.

\section{Conclusion and future work}
We have shown that the GPU can be used to improve the performance of a parallel SAT solver. Given a large set of assignment (up to 1024), and millions of clauses, the GPU is very efficient in noticing a posteriori which clause would have been useful to have on these assignments. This allows CPU solver threads to efficiently import only clauses that would have been useful to have.

The GPU reads the assignments from global memory. We could probably make the GPU much faster by reading some of them from shared memory. If, inside a GPU block, all the threads were to look at clauses that are very close to each other (i.e. similar variables): by putting the values of some of these variables in shared memory, we could probably improve GPU performance.

In our algorithm: testing an aggregate over 32 assignments is enough in the vast majority of cases to tell that a clause does not trigger. For this reason, we could try to compute the aggregate of 32 aggregates of up to 32 assignments. When the current algorithm handles up to 32 x 32 assignments, that would allow us to handle 32 x 32 x 32 assignments.

Our solver is based on glucose-syrup 4.1, and the performance comparison has been made against that solver. We should try to add our new clause exchange policy on a newer and more performant SAT solver, and see how it compares.

\bibliographystyle{plain}
\bibliography{paper}

\end{document}